\documentclass[conference]{IEEEtran}
\usepackage{times}

\let\labelindent\relax
\usepackage{enumitem}
\setitemize{noitemsep,topsep=0pt,parsep=0pt,partopsep=0pt}

\usepackage[numbers]{natbib}
\usepackage{multicol}
\usepackage[bookmarks=true]{hyperref}
\usepackage{setspace}
\usepackage[disable]{todonotes}

\newcommand{\pk}[1]{\todo[inline,color=blue!10]{pk: #1}}

\usepackage[utf8]{inputenc} %
\usepackage[T1]{fontenc}    %
\usepackage{hyperref}       %
\usepackage{url}            %
\usepackage{booktabs}       %
\usepackage{amsfonts}       %
\usepackage{nicefrac}       %
\usepackage{microtype}      %
\usepackage{cleveref}  %

\usepackage{graphicx} %
\usepackage[caption=false]{subfig}  %
   \graphicspath{{figs/}}
   \DeclareGraphicsExtensions{.pdf,.jpg,.png,.eps}

\usepackage{siunitx}
\usepackage{multirow}

\usepackage{floatrow}
\newfloatcommand{capbtabbox}{table}[][\FBwidth]
\usepackage{amsmath,amssymb}
\usepackage{xspace}
\usepackage{balance}  %

\newcolumntype{L}{S[table-format=2.1]@{\hskip 10pt}}
\newcolumntype{R}{S[table-format=2.1]}
\newcolumntype{?}{!{\vrule}}
\def\abovestrut#1{\rule[0in]{0in}{#1}\ignorespaces}

\newcommand{\myparagraph}[1]{\textbf{#1}~} %

\newcommand{\mytimes}{\medmuskip=0mu\times}

\newcommand{\secref}[1]{Section~\ref{#1}}
\renewcommand{\eqref}[1]{(\ref{#1})}
\newcommand{\figref}[1]{Fig.~\ref{#1}}

\newcommand{\tabref}[1]{Table~\ref{#1}}

\newcommand{\ie}{\textit{}{i.e.}}
\newcommand{\eg}{\textit{e.g.}}
\newcommand{\etc}{\textit{etc.}}

\usepackage{color}
\definecolor{fullred}{rgb}{0.95,.0,.1}
\newcounter{cmt}

\newcommand{\prob}{\ensuremath{p}}

\newcommand{\rew}{\ensuremath{\mathcal{J}}}

\begin{document}

\title{Differentiable Algorithm Networks for\\ Composable Robot Learning}

\author{\authorblockN{
Peter Karkus$^{1,2}$,
Xiao Ma$^{1}$,
David Hsu$^{1}$, 
Leslie Pack Kaelbling$^{2}$,
Wee Sun Lee$^{1}$ and
Tom{\'a}s Lozano-P{\'e}rez$^{2}$}
\authorblockA{$^1$National University of Singapore, $^2$Massachusetts Institute of Technology}
\authorblockA{\texttt{karkus@comp.nus.edu.sg}}
}

\maketitle

\begin{abstract}
This paper introduces the \textit{Differentiable Algorithm Network}~(DAN), a composable architecture for robot learning systems. 
A DAN is composed of  neural network modules, each  encoding  a differentiable robot algorithm and  an associated model; and it is trained end-to-end from data. DAN combines  the strengths of model-driven modular system design and data-driven end-to-end learning. The algorithms and  models act as structural assumptions  to reduce the data requirements for learning;   end-to-end learning allows the modules to adapt to one another and compensate for imperfect models and algorithms, in order to achieve the best overall system  performance.
We illustrate the DAN methodology through a case study on a simulated robot system, which learns to navigate in complex 3-D environments with only local visual observations and an image of a partially correct 2-D floor map.
\end{abstract}

\IEEEpeerreviewmaketitle

\section{Introduction}

There is an essential tension between the model-based and the model-free approaches to robot system design. Robotics research has provided a wealth of powerful models for capabilities including perception, state estimation, planning, and control. Put together, they form the basis of  many successful robot systems, from Shakey~\cite{nilsson1984shakey} to Stanley~\cite{thrun2006stanley}. 
At the same time, the data-driven, model-free approach, particularly, deep learning, has recently produced exciting results in areas such as vision and object manipulation (see, \eg, ~\cite{ And18, krizhevsky2012imagenet,mahler2017dex}), tasks in which the model-based approach faces much difficulty despite decades of research.  Can we reconcile the seemingly conflicting assumptions of  model-based and model-free approaches and integrate the two into a single unified framework?

While the two approaches appear antithetical, they in fact focus on different aspects of robot system design. 
The   model-based approach focuses on the structured, modular representation. We decompose a robotic task into well-understood, interpretable sub-tasks. For each sub-task, we construct a model manually from prior knowledge or learn it from data. We then develop algorithms to infer solutions given the model. Finally, we compose the  components into an overall system.   The performance of the system may be suboptimal because of imperfect models, approximate algorithms, or a poorly chosen decomposition. 
In contrast, the model-free approach relies on end-to-end training with powerful function approximators. We approximate the robot policy or controller with  a rich parameterized function; we learn all parameters jointly from data, and optimize for the overall objective. A key issue here is to choose the approximating function class. We need the right prior, or \textit{bias}, to moderate the data requirements for learning.
We want to combine the strengths of the model-based and model-free approaches
by performing end-to-end learning over a structured, modular system
representation. Our objectives are (i) to compensate for imperfections in
models, algorithms, and decomposition and (ii) achieve strong performance with
limited training data.

\begin{figure*}[!t]
  \centering
  \hfill
  \subfloat[][Model-based]{ 
  \includegraphics[width=0.3\textwidth]{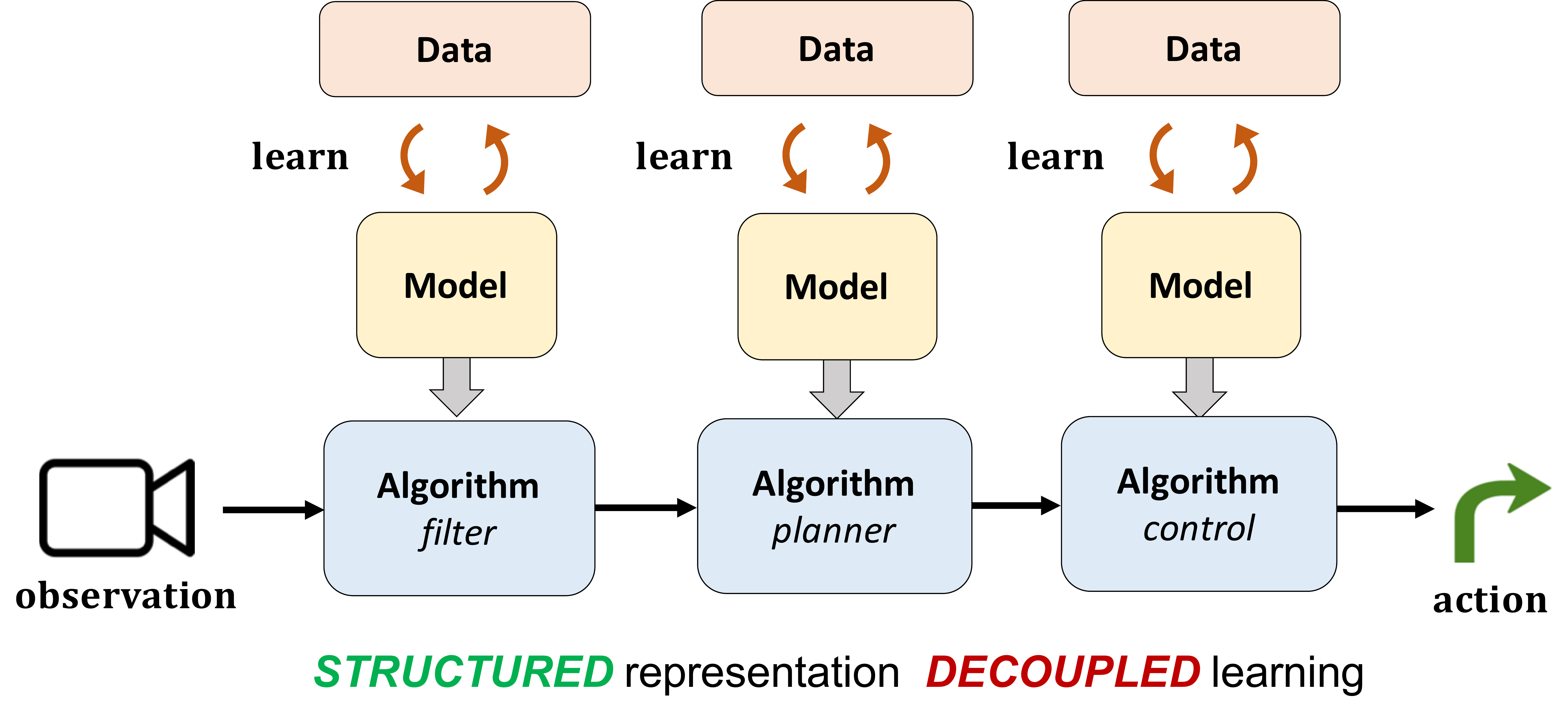}\hspace{0.3cm}\label{fig:model_based}}
  \hfill
  \subfloat[][DAN]{
  \includegraphics[width=0.3\textwidth]{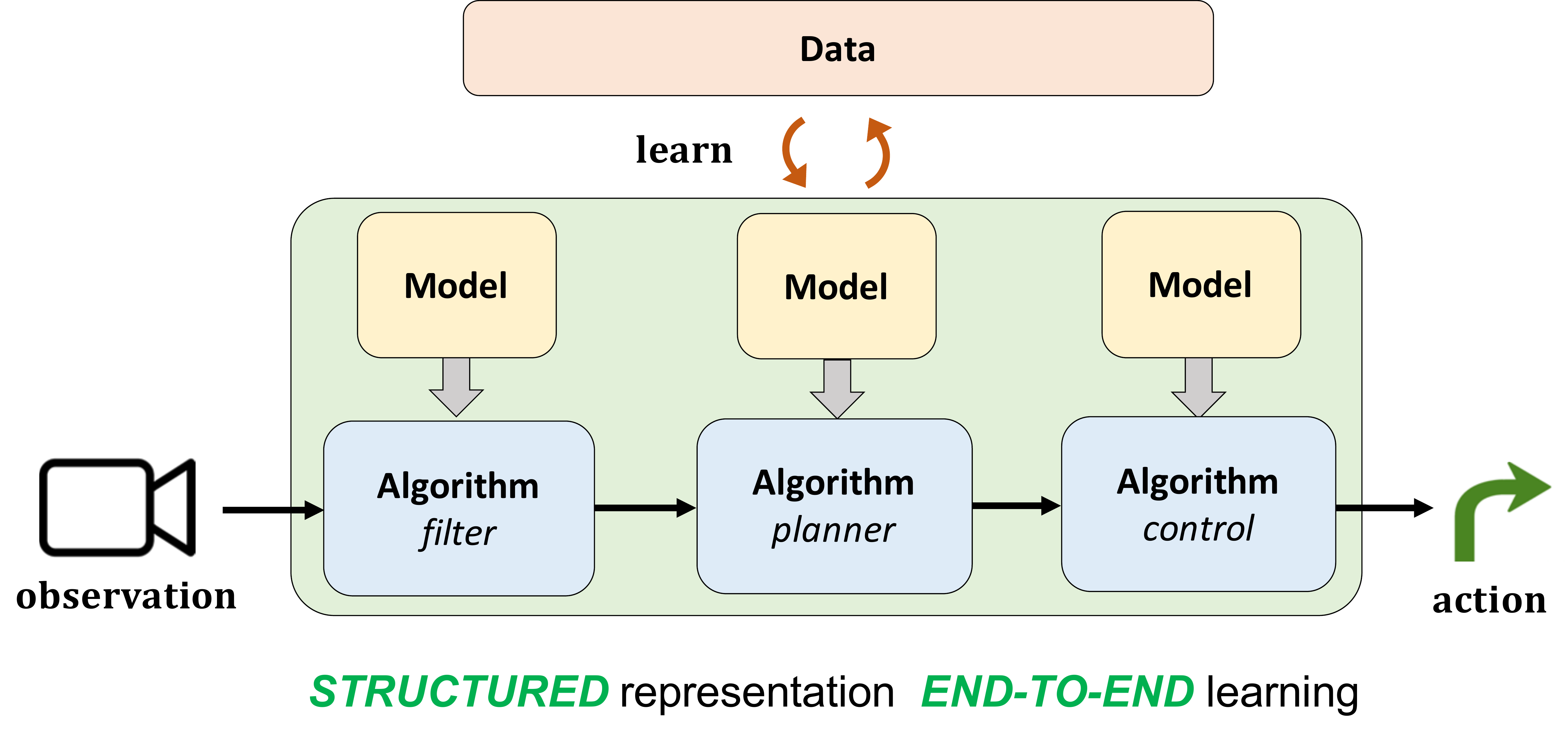}\hspace{0.3cm}\label{fig:dan}}
  \hfill
  \subfloat[][Model-free]{
  \includegraphics[width=0.3\textwidth]{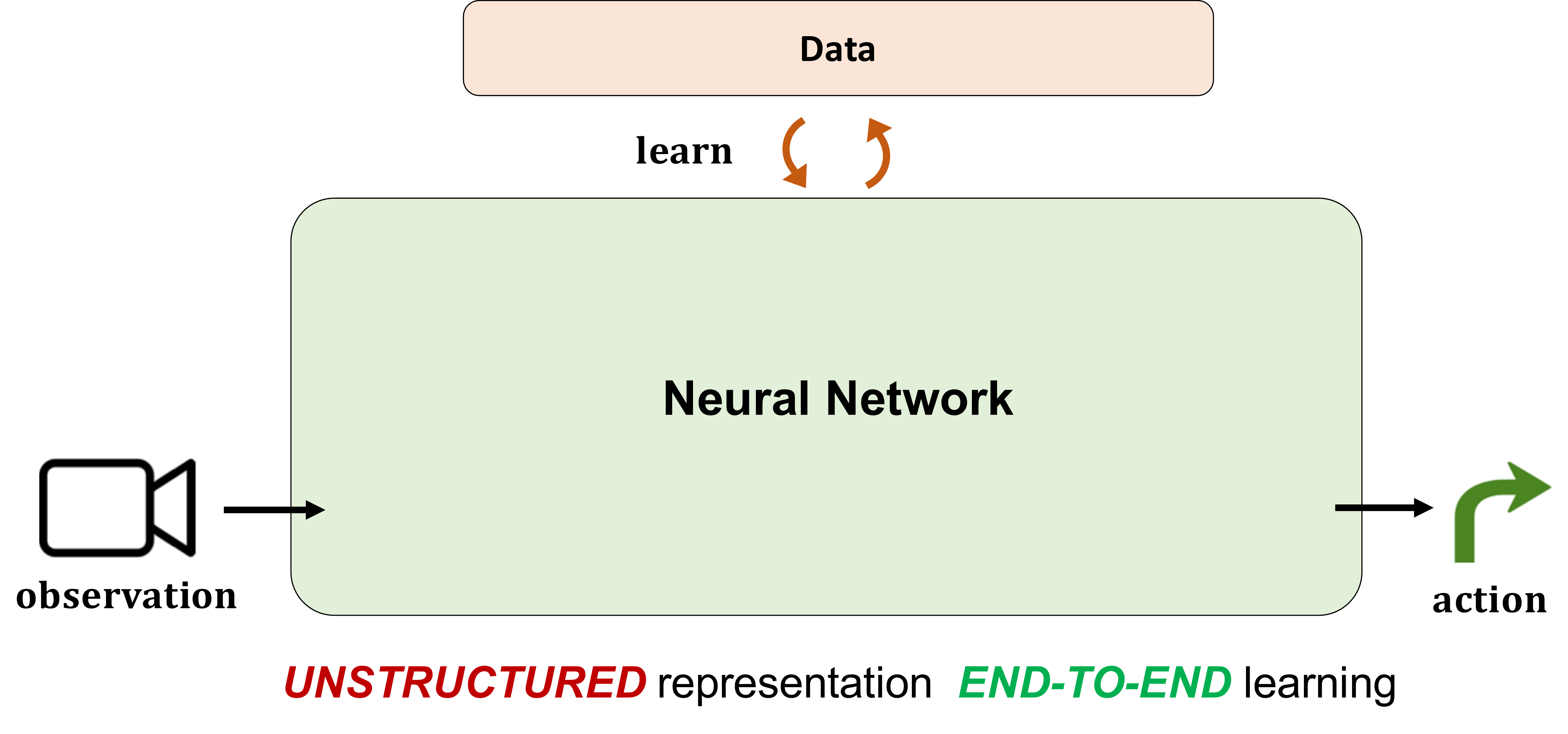}\hspace{0.3cm}\label{fig:model_free}
  \hfill}%
  \caption{Three general architectures of robot learning systems. 
  (a) Model-based: structured representation, decoupled learning. 
(b) DAN: structured representation \textit{and} end-to-end training.  (c) Model-free: unstructured representation, end-to-end learning.\vspace{-6pt}}
\label{fig:approaches}
\end{figure*}

To this end, we introduce the \textit{Differentiable Algorithm Network} (DAN), a composable architecture for robot learning systems.  A DAN is composed of neural network modules encoding differentiable  robot  algorithms  and  associated  models; and it is trained  end-to-end  from  data. The network architecture is constrained by the algorithms, which act as a structured prior.
The network parameters encode the model parameters and tunable algorithm parameters. In contrast with the conventional model-based approach, which learns a model to match the underlying system dynamics, DAN trains the network
end-to-end to optimize the overall task objective by allowing  network modules, including both  models and  algorithms,  to  adapt and compensate for each others' imperfections. 
The idea of embedding algorithms in neural networks has appeared many times before (\eg,  \cite{amos2018differentiable, amos2017optnet, donti2017task, farquhar2017treeqn, guez2018learning, haarnoja2016backprop, jonschkowski2016, jonschkowski2018differentiable, karkus2017qmdp, karkus2018particle, kloss2019on, pereira2018mpc, racaniere2017imagination, srinivas2018universal, oh2017value, okada2017path, tamar2016value}).
Recent studies suggest that even biological nervous systems encode model and algorithm structures, \eg, the Bloom filter~\cite{bloom1970space} 
in the fruit fly~\cite{dasgupta2018neural}. However, earlier work all focuses on individual algorithm modules rather than composing them for an overall system.

DAN provides a general methodology for composing flexible, robust robot learning systems. We illustrate this methodology through a  case study of building a simulated robot system that navigates in  complex 3-D environments with only local visual observations and an image of a partially correct 2-D floor map.   The architecture of this system includes DAN modules for visual perception, state filtering, planning, and local  control.   We experiment with several  learning strategies  and use the results to highlight three main advantages of the DAN methodology.

\begin{enumerate}

\item \textbf{Robustness against imperfect models and algorithms}. 
Strong priors, in this case, model-based algorithms, speed up learning. However, poorly chosen priors may ultimately limit the performance of the learned system.  DAN learning systems are robust. Even when the underlying algorithm of a DAN  makes  simplifying assumptions that are not fully satisfied, such as the Markov property or perfect observations, the overall system can learn a model that compensates for the mismatch between the algorithm's assumptions and the  physical reality.

\item \textbf{Robustness against imperfect system decomposition}. 
The venerable modularity principle of system design  requires us to specify well-defined interfaces to decompose a system into self-contained components. 
Imperfect independence assumptions may result in suboptimal interface choices.
By relaxing the interfaces between modules through learning, DAN improves overall system performance, sometimes significantly.

\item \textbf{Flexible  module representation}. Neural networks, treated as computational graphs, provide a rich, flexible representation. While some system modules are naturally represented as an algorithm together with a model, others are more easily represented as a linear feedback controller, finite-state machine, or a recurrent neural network. DAN provides a uniform representation for  them all and a standard interface for composing them.

\end{enumerate}

\section{Related work}
\label{sec:related}
The DAN compensates for imperfect model assumptions in a robotic system through end-to-end training. 
Imperfect models are common  in robotic systems. They  have been studied widely in various contexts, \eg,  to  account for  uncertainty in the planning model~\cite{deisenroth2011pilco, doshi2012reinforcement, jiang2015dependence, van2011lqg, talvitie2014model} or to directly
 learn strategies that are
 robust against imperfections in  environment models~\cite{cutler2015real, ajay2018augmenting},  policies~\cite{johannink2018residual, silver2018residual}, or approximate algorithms~\cite{racaniere2017imagination}.
Unlike the earlier work, the DAN commits to the algorithm choices, but adapts the models to compensate for imperfections through end-to-end training from data.

Learning models adapted to a task or an algorithm has been also explored in the past. In the context of MDPs, \citet{farahmand2018iterative} learns transition models adapted to value iteration to compensate for a mis-specified model class. 
In the context of manipulation, \citet{agrawal2016learning} learns inaccurate, ``intuitive'' models that still allow good task performance. %
In the context of control, \citet{bansal2017goal} applies Bayesian optimization to learn a transition model directly for the policy. 
DAN builds on a differentiable representation of algorithms, which scales to learning complex models in large, modular systems jointly.

The idea of encoding algorithms in neural networks has been proposed in various contexts, including state estimation~\cite{haarnoja2016backprop, jonschkowski2016, jonschkowski2018differentiable, karkus2018particle}, planning~\cite{farquhar2017treeqn, guez2018learning, karkus2017qmdp, oh2017value, tamar2016value} and control~\cite{amos2018differentiable, donti2017task, okada2017path, pereira2018mpc}. DAN generalizes these ideas: it encodes the entire solution structure of a system including multiple model and algorithm components. 
In this paper we investigate the opportunities the DAN approach opens up in robot system design,
where incorrect modelling assumptions are compensated for by using end-to-end learning from data. While prior work shows potential in this direction~\cite{karkus2017qmdp}, general DANs with multiple components have not yet been explored.

We study DAN in a visual localization and navigation domain, a challenging robotic task of real importance~\cite{desouza2002vision, anderson2018evaluation}.
While our primary objective is to illustrate the DAN approach,
our work can also be seen as a contribution to the state of the art in constructing robust visual navigation systems. Our problem setting features important challenges that are generally not present in related work, \eg, \cite{mirowski2016learning, zhu2017target}. Specifically, the goal is specified on a map, which {\em requires} localizing and planning with respect to the map. The combination of uncertain location, partial map, and visual input makes this a difficult partially observable decision-making problem, which is challenging for both model-based and model-free approaches.

\section{Differentiable algorithm networks}

\subsection{General Architectures for Robot Learning Systems}

There is a broad spectrum of different architectures for robot learning systems, with the purely model-based approach and the purely model-free approach sitting at the extremes (\figref{fig:approaches}). DAN aims to combine the strengths of both.

The model-based modular system design decomposes a system into well-understood components, such as filters, planners, and controllers, with well-defined interfaces between them (\figref{fig:model_based}).
There is a clear separation between \textrm{models} and \textrm{algorithms}. Each model is designed or trained independently %
to match the underlying physical reality.  
For example, we may learn a probabilistic  state-transition model $\prob(s' | s, a)$ for state~$s$, action~$a$, and next state~$s'$, given supervised training data  $(s_i, a_i, s_{i+1}), i=0, 1, 2, \ldots$\,, or learn an observation model $\prob(o | s)$ for state $s$ and observation $o$, given the data $(s_i, o_i), i=0,1,2, \ldots$\,.
We learn these models by maximizing the model likelihood of the training data. 
The model-based approach relies  on  well-understood, generally correct ``independence'' assumptions to decompose a system into modules and interfaces. It has produced many successful robot systems.

However, some robot tasks are poorly understood. Identifying the right assumptions for modeling or decomposition is difficult. Consider,  for example, folding clothes.  We may not know a good representation of latent states. Standard modeling assumptions, such as Markovian state transitions or Gaussian noise, may break down. Large observation spaces, such as camera images, make learning a complete probabilistic distributional model infeasible. In reaction to these difficulties,  the model-free, end-to-end approach abandons models completely; instead, it exploits the strong approximation capabilities of a general function approximator, such as a deep neural network, and uses large amounts of data to train end-to-end on the task of interest (\figref{fig:model_free}). %
For example, it may train a neural network policy that directly maps camera images to robot actions for manipulating clothes. 
The lack of assumptions, however, often comes at the cost of large amounts of training data, reflecting the fundamental trade-off between model assumptions and data.

DAN fuses the model-based and model-free approaches (\figref{fig:dan}). Like the model-based approach, it exploits domain knowledge to design the overall system structure: it embeds model-based algorithms for  filtering, planning, and control in a neural network and also maintains the separation between  models and algorithms. These structural assumptions provide a strong and useful bias, which reduces the required amount of training data.
At the same time, like the model-free approach, DAN %
trains the entire system  end-to-end, thus 
allowing modules to modify themselves cooperatively to optimize overall task performance and to compensate for any incorrect model assumptions. Further, the uniform neural network representation makes it easy to mix model-based and model-free elements within a single system. For example, we can replace a model-predictive control module with an LSTM~\cite{hochreiter1997long} or \textit{vice versa}. \pk{skip last two/one sentences if need space}

\subsection{Differentiable Algorithms}

DAN is based on the idea of \textit{differentiable algorithms}. 
 We view a robot  system as a policy $\pi$ that maps observation histories to robot actions. Ultimately we desire a system that performs well according to a suitable metric $\rew({\pi})$, \eg, expected total reward. We may obtain $\pi$ by applying an algorithm $\mathcal{A}$ to a model $M$, so that $\pi= \mathcal{A}(M)$. %
The model $M$ often takes on a parametric form $M(\theta)$ and is learned from data.
To construct a DAN, we start with the same conceptual structure, but design a representation of $M(\theta)$ and a neural network function ${F}$, so that ${F}(\theta) \approx \mathcal{A}\bigl(M(\theta)\bigr)$. 
Most importantly, $F(\theta)$ is \emph{differentiable} with respect to $\theta$.  

What is the benefit of a differentiable representation of $F(\theta)$?  The conventional model learning objective is a form of predictive likelihood  $\ell\bigl(M(\theta)\bigr)$, which is independent of the algorithm $\mathcal A$. It is  only indirectly connected to the end objective $\rew(\pi)$ through $\pi = \mathcal{A}\bigl(M(\theta)\bigr)$. 
In contrast, DAN learns by directly optimizing  $\rew(\pi) = \rew\bigl(F(\theta)\bigr)$. This end-to-end optimization is generally difficult, because it involves $\mathcal A$. The differentiable representation of $F(\theta)$ allows for efficient first-order methods, \eg, gradient descent, which  back-propagate gradients through the steps of $\mathcal A$ encoded in $F(\theta)$.

The premise of the DAN methodology is that many key robot algorithms admit  differentiable representations. Prior work has addressed this important question for filtering, planning, and control (see Section~\ref{sec:related}).

A differentiable representation  is straightforward if the algorithm contains only differentiable operations. For example, the histogram filter only uses matrix multiplications and summations, which are clearly differentiable. Other algorithms involve non-differentiable operations, such as discrete maximization, sampling, and indexing. 
One strategy is to replace non-differentiable operations by a differentiable approximation, \eg, replace max by soft-max, sampling by soft-sampling~\cite{karkus2018particle}, indexing by soft-indexing~\cite{karkus2017qmdp}. A drawback of these new operations is a possible reduction of algorithmic efficiency. Another strategy is to keep non-differentiable nodes in the computation graph, and approximate the gradients, e.g. through sampling methods. An example %
is the implementation of Monte-Carlo tree search in~\cite{guez2018learning}. \citet{schulman2015gradient} provides a generic framework for optimizing computation graphs with non-differentiable nodes, although gradient estimates can have high variance. Generic variance-reduction techniques are actively being investigated~\cite{weber2019credit}. \pk{remove prev sentence for space} 
Finally, a fundamentally different strategy is to derive analytic gradients around a fixed point output of the algorithm. The idea has been explored in the context of control algorithms~\cite{amos2018differentiable} and for generic optimization~\cite{amos2017optnet, chen2018neural}. \pk{shorten for space}
When applicable, the analytic approach is appealing for its computational efficiency. 

\subsection{Compensating for Approximations: Illustrative Examples}

We begin with two simple %
examples that illustrate DAN's ability to compensate for imperfect models and algorithms.

\begin{figure}[!t]
  \centering
  \subfloat[][Puddle-MDP]{ 
  \includegraphics[height=2.1cm]{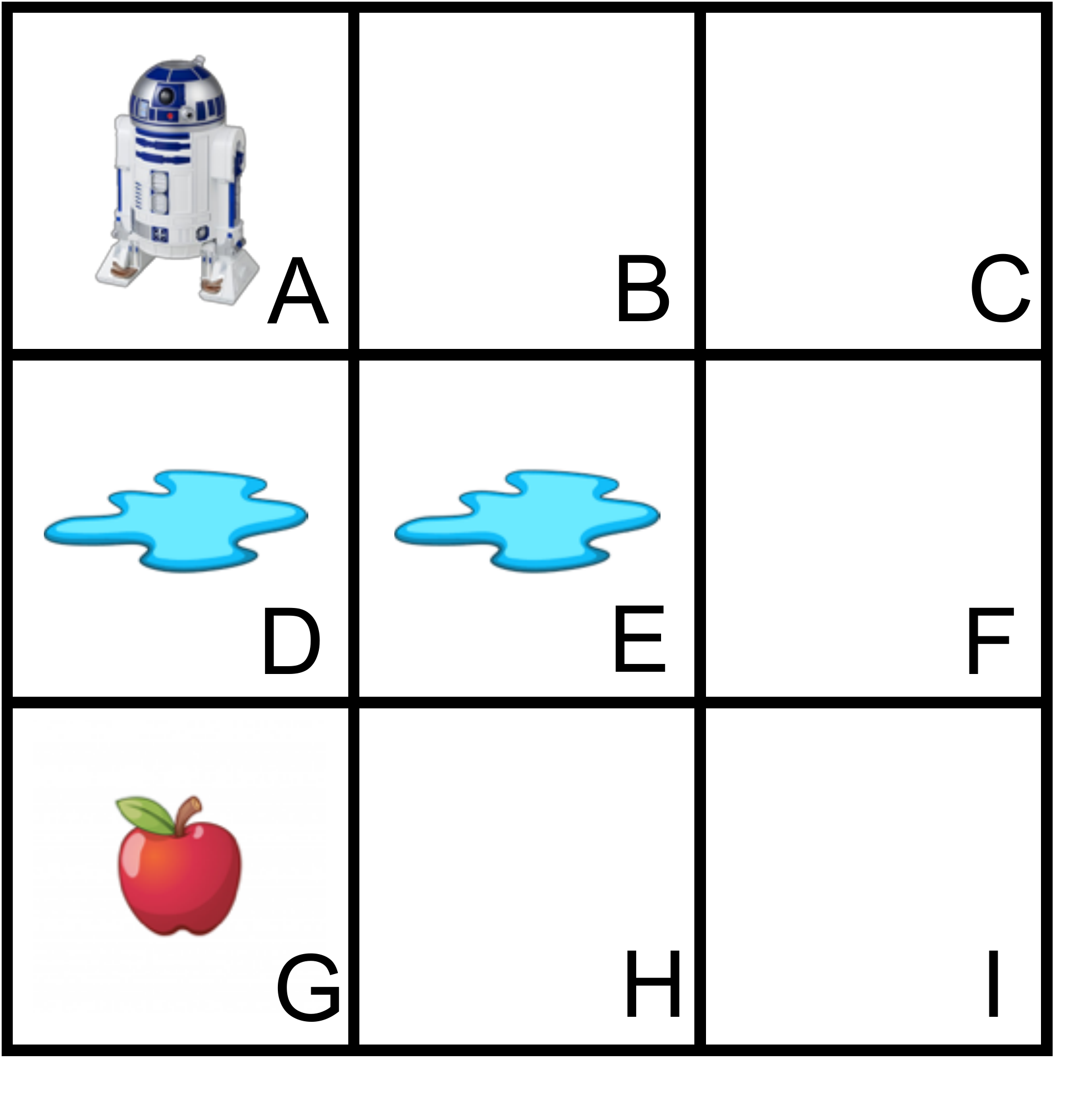}\label{fig:puddle_mdp}}
  \hspace{0.3cm}
  \subfloat[][Puddle-POMDP]{
  \qquad\includegraphics[height=1.45cm]{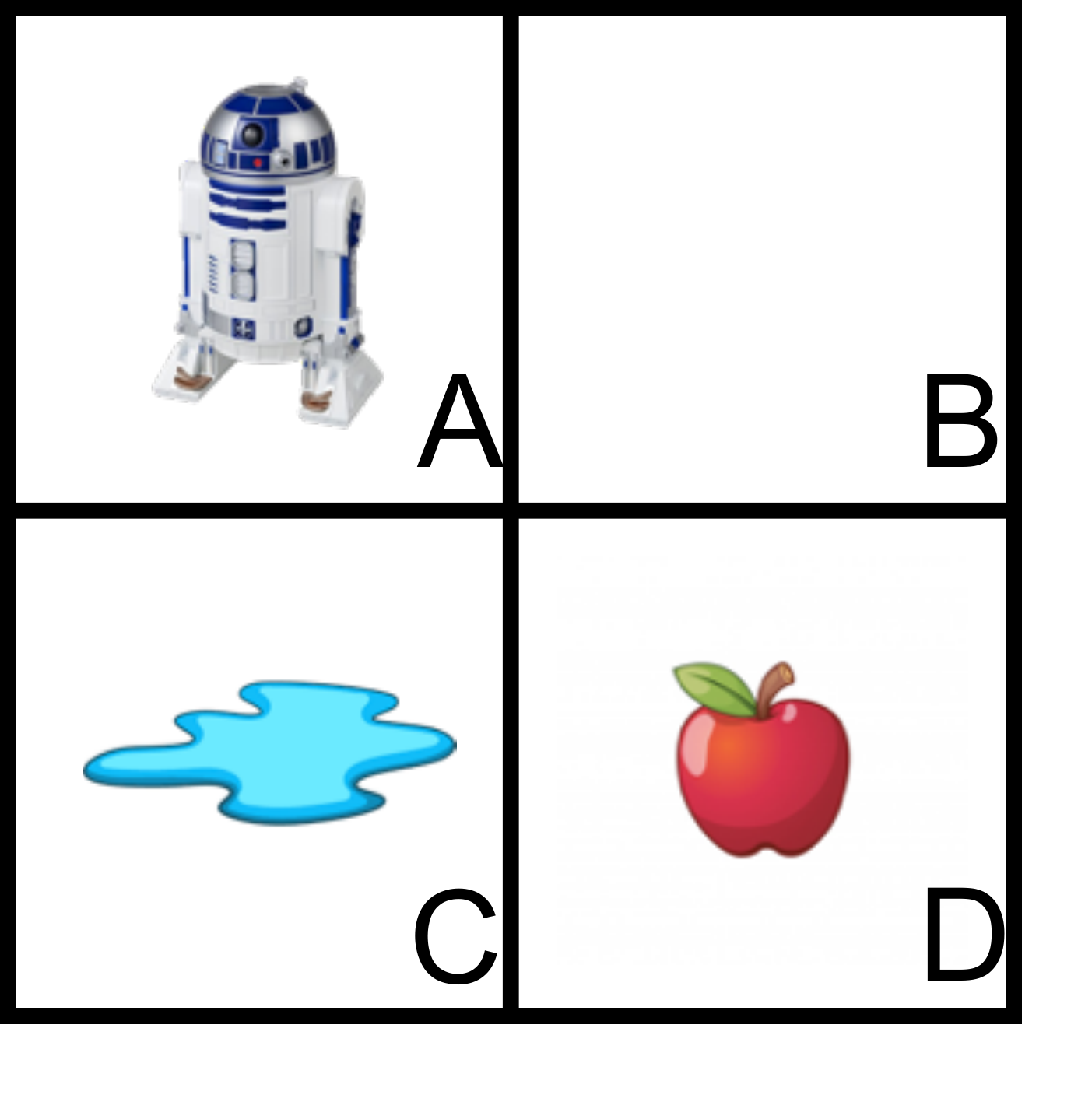}\qquad\label{fig:puddle_pomdp} }%
  \caption{Simple illustrative problems.}
\label{fig:puddle}
\end{figure}

\myparagraph{Puddle-MDP} 
Consider a tiny grid navigation problem, Puddle-MDP, in \figref{fig:puddle_mdp}. Actions are to move right, left, up, or down.  Reaching the apple ($G$) or a puddle ($D$ and $E$) yields positive and negative rewards, respectively.  Starting in state $A$, the best strategy is to move along path $ABCFIHG$, reaching the apple while avoiding puddles.  
In a model-based approach we may learn transition and reward models, and plan with the value iteration algorithm. If the planning horizon is 6 or greater, we get the optimal policy. 
However, what if the planning horizon is 4? 
Value iteration with the perfectly accurate transition and reward models cannot find a way around the puddles. We train a DAN, encoding the horizon-4  value iteration algorithm, and get a surprising result: the learned transition model encodes actions as if they were ``macro-actions,'' predicting that moving right from state $A$ will land in state $C$, \etc.
\pk{evidence}
While this model is predictively incorrect, it allows value iteration to find the optimal path around the puddles even with a horizon of 4. Similar algorithm approximations are common. As illustrated by this example, DAN may compensate for an approximate algorithm  by training models end-to-end.

\myparagraph{Puddle-POMDP}  We also consider a related problem, shown in~\figref{fig:puddle_pomdp}. The robot starts in state $A$, but it does not observe the state afterwards; and actions may fail ($p_\textrm{fail}=0.4$), in which case the robot remains in place.  The optimal solution is non-trivial. If the robot moves right and down, it will often end up in the puddle due to the transition noise. The optimal policy is to take the right action multiple times, until the state uncertainty is sufficiently reduced, and only then move down.  Again, when we learn the ``correct'' transition and observation models, and use a POMDP-solution method, the optimal strategy is found.  POMDP solvers are computationally expensive. 
Robot systems often decouple state-estimation, and plan with the most likely state. In this problem, having the ``correct'' models, the decoupled strategy performs poorly: after moving one step to the right, the most likely state is $B$, in which the robot should move down---but there is a substantial chance the robot will have stayed in $A$ and moved into the puddle.
We encode the same system, state estimator and planner, in a DAN, and train the models end-to-end, to generate good behaviour instead of accurate predictions. The optimal behavior is recovered. How? The ``failure'' probability is increased over $0.5$,
\pk{evidence}
so that after moving to the right, the most likely state is still $A$, which causes additional rightward motions before moving down.  Similar situations occur in practice: for example, Monte-Carlo localization is known to work better when the transition noise is overestimated \cite[p.~118]{thrun2005probabilistic}, \cite{jonschkowski2018differentiable}. Through end-to-end training, DAN may learn similar strategies for compensating modelling approximations. \pk{cut down for space}

\section{Case study: Learning visual navigation under uncertainty}
We investigate the DAN approach in a simulated visual navigation domain. The domain highlights important challenges of robot decision making: acting under state uncertainty, environment uncertainty, and processing rich sensory input. Because of the combination of these challenges, an adequate model-based system design is not immediately available---modelling approximations are necessary. %
Our case study reveals various ways DAN training may compensate for modelling approximations, even in a moderately large modular system, enabling substantially improved performance.

\subsection{Domain Description}

\begin{figure}[t]
\includegraphics[width=0.84\linewidth]{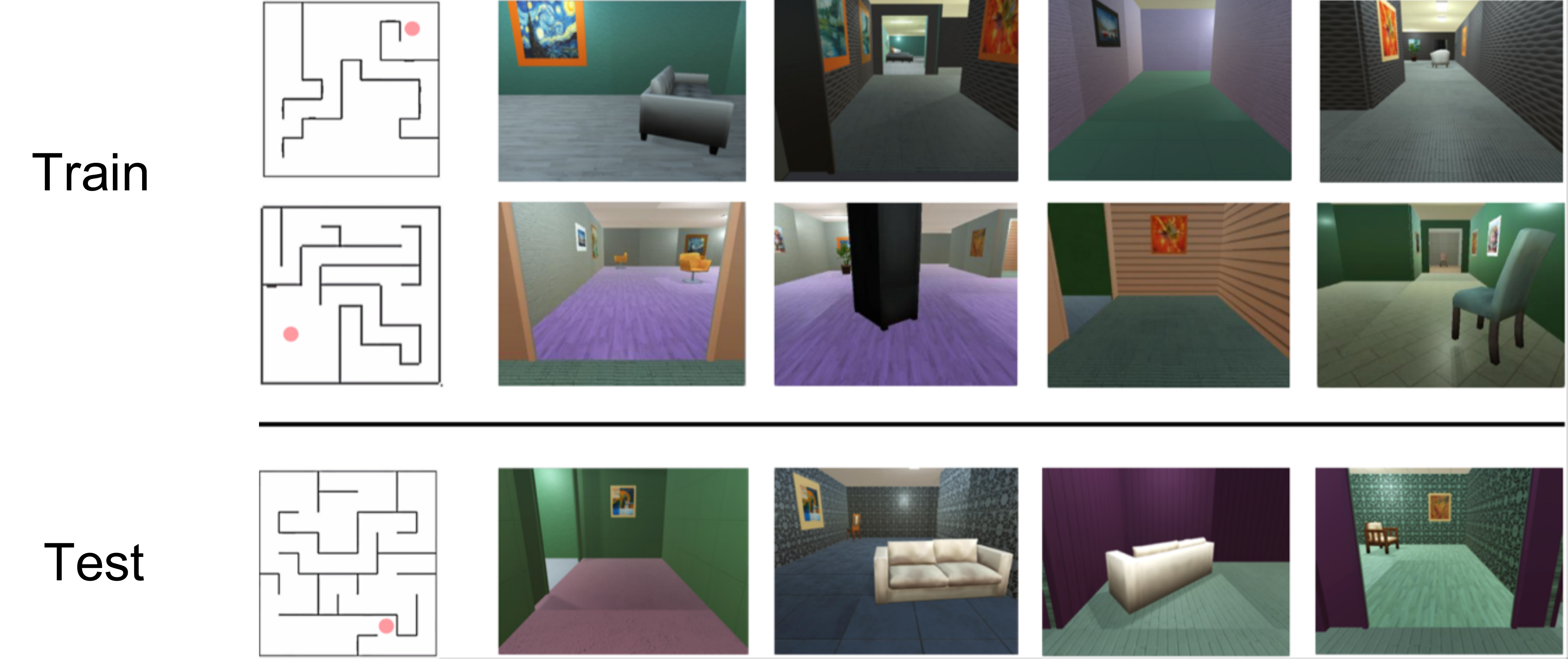}
\caption{3D simulation environments for navigation. The location of the robot is not observed. Maps do not indicate objects to be avoided. } 
\label{fig:examples}
\end{figure}

A robot is tasked to navigate to a goal in a previously unseen, visually rich 3D environment, using a 2D floor map, and images from an onboard camera. Each environment has different layout and visual appearance.  Examples are shown in \figref{fig:examples}. The domain involves challenging partial observability: the location of the robot is unknown, and the environment is uncertain, \ie, the map indicates walls, but not other objects, like furniture. Since the goal is specified on the map, the robot must localize, which involves matching features from rich camera images and the 2D map. %
It must then find a path to the goal, potentially far from its current location, and navigate while detecting and avoiding unknown objects in the environment.  We have access to a set of training environments for learning. After learning, the system is evaluated in new, previously unseen environments.

We develop a custom-built simulator with the Unity 3D Engine~\cite{unity3d} for controlled experiments. 
The simulator captures the critical challenges in terms of uncertainties and rich visual appearance; however, it does not aim to simulate all aspects of the real-world, such as dynamics. %
The simulator generates randomized 3-D environments. First, a random $19\mytimes19$ grid maze is generated that defines the placement of walls in the 3-D environment. Second, additional objects are placed at random locations, but without fully blocking passages. Objects are chosen from a set of 23 common household furniture: chairs, tables, beds, \etc{} %
The visual appearance is  randomized, including pictures on the walls, textures, colors and lighting.
The motion of the robot is simplified: robot states are discrete 2-D coordinates with orientation, and transitions are deterministic. There are a total of 1444 states corresponding to the $19\mytimes 19$ grid and 4 orientations. The robot does not observe the state. It is given a set of possible initial states: in some episodes this includes all states, in some episodes a random subset of all states. 
Collisions with objects are evaluated based on the robot bounding box of size $0.8\mytimes 0.8$ grid cells. In the event of a collision the state remains unchanged. The actions are: {\em move forward}, {\em turn left}, {\em turn right}, {\em stay}. Rewards are $+20$ for reaching the goal; $-10$ for a collision; $-0.1$ for every other action. 
The discount factor is $0.99$.

We consider variants of the domain with increasing difficulty, as shown in~\tabref{tab:task_variants}.
In Task~A, the robot location is directly observed, and the map given to the robot fully describes the environment, including walls and furniture.
In Task~B, the location is not observed. Instead, the robot receives local (but  noise-free) observations indicating the presence of walls or objects in the 3 grid squares in front of the robot.
In Task~C, the noise-free binary observation vectors are replaced with images from an onboard camera. %
Finally, Task~D is the full domain that involves vision, uncertain location and uncertain  environment. The map given to the robot only partially describes the environment: it indicates walls but not the furniture. %

\subsection{Architecture Description}\label{sec:architectures}

\begin{figure}[t]
\includegraphics[width=0.95\textwidth]{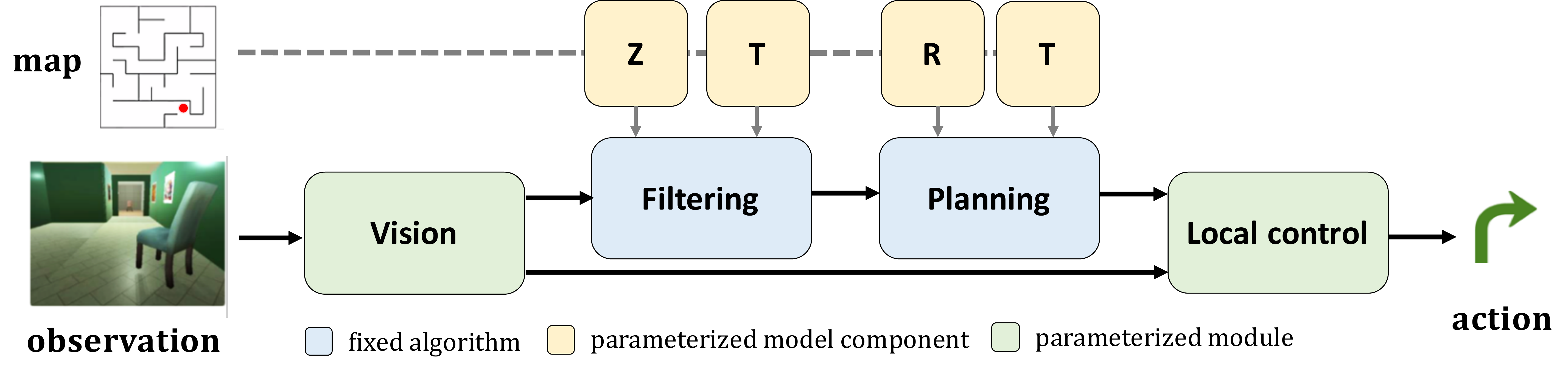}
\caption{System architecture for visual navigation.}
\label{fig:navigation_architecture}
\end{figure}

We experiment with different architectures for systems that learn in this domain. The systems are built by composing up to four generic modules: vision, filtering, planning, and local control~(\figref{fig:navigation_architecture}). We explore different implementations and training strategies for each module and combinations of modules.  The different versions of the modules appear as entries in our results table (\tabref{tab:main_results}). Details are in the appendix.

\begin{table}[t]
\vspace{-8pt}  %
\scalebox{0.68}{
\begin{tabular}{lcccc}
 \toprule
    & \textbf{Task A} & \textbf{Task B} & \textbf{Task C} & \textbf{Task D}\\
 \midrule[0.08em] %
 Map input &  full  & full & full & without furniture \\
 Observation input & location & vector & image & image \\
 \midrule
 Uncertain robot location       &  & x & x  & x \\
 Visual input                        &  & & x  & x \\
 Partially correct map     &  & & & x  \\
 \bottomrule
\end{tabular}
}
{\caption{Task variants and challenges involved}\label{tab:task_variants}\vspace{-3pt}} %
\end{table}

\myparagraph{Vision}  
The vision module takes high-dimensional, $80\mytimes40\mytimes3$ images as inputs, and outputs a low-dimensional vector representation. We consider two implementations (denoted CNN and CNN-f)  to explore the effect of relaxing the interfaces between modules.  CNN is a convolutional network~\cite{lecun1998gradient} with a length-3 binary vector output that indicates the presence of walls and objects in the 3 grid squares in front of the robot; this output is compatible with the observations provided as input in Task B. CNN-f is a similar network that outputs length-16 embedding with no pre-specified semantics.

In Task D, where the map is partial, we use two vision modules: one for estimating location in the map (filtering) and one for local control. The output observation vector in the first case is trained to only indicate walls. In the second case it is expected to indicate all obstacles, both walls and objects.  The neural network weights in the two modules are shared except for the last fully-connected layer.

\myparagraph{Filtering}  
The filtering module is a histogram (Bayes) filter \cite{thrun2005probabilistic, jonschkowski2016} that maintains a \emph{belief}, a probability distribution over the states of the robot, which will be the input to the planner.  The default implementation (HF-bel) takes a length-3 observation vector, the action, and previous belief as inputs, and outputs an updated belief (probabilities over states).  We also experiment with a version of the filter module (HF-ml) that outputs a one-hot encoding of the most-likely state.  In the DAN setting, we use the notation HF to indicate that
the semantics of filter output is not enforced. 
A version that has the richer length-16 feature vectors as input is denoted as f-HF.

The histogram filter uses two parameterized models: an observation model, $Z(o_t| s_t)$, that defines the probability of observations given the state; and a transition model $T(s_{t+1} | s_t, a_t)$ that defines the probability of next states given the current state and an action.  We represent these models by small neural networks conditioned on the map. Specifically, the observation model combines features from the $19 \mytimes 19$ map and the observation vector, and outputs a $1444$-dimensional vector, estimates of the observation likelihood for each state.

The transition model defines $3\mytimes 3 \mytimes 4$ local transition probabilities for each state and action pair. We consider two implementations: a heterogeneous model, where local transition probabilities are estimated from the map using a convolution layer; and a homogeneous model that is independent of the map and the states. The parameters in the latter case are the local transition probabilities for each action; this is used in the DAN setting. %
In the independently learned settings, the transitions are heterogeneous, giving more accurate models.

\myparagraph{Planning} 
We use two planning algorithms: value iteration (VI), a simple method for solving MDPs~\cite{bellman2013dynamic, tamar2016value}, and SARSOP, a state-of-the-art POMDP planner~\cite{kurniawati2008sarsop}.  Both planners take in a belief vector and output an action.  The planners require transition and reward models. Transition models are identical to the histogram filter, alternatively homogeneous or heterogeneous. The reward model, $R(s, a)$, defines rewards for each state action pair. We assume rewards are unknown. The reward model is learned using a two-layer CNN that takes in the map and outputs a $1444$-dimensional reward vector for each action. %

VI planning is done by applying Bellman updates for $H$ (horizon) iterations, which provide approximate Q values for each state-action pair. The length-4 action-value outputs are obtained by weighting the Q values for each action with the belief vector. When the belief vector is a one-hot encoded state, this corresponds to taking the Q values for the current state. When the belief vector is a distribution, this strategy is known as the QMDP approximation~\cite{littman1995learning, karkus2017qmdp}. 

We explore a number of variations on VI: %
long %
and short horizon, and heterogeneous %
and homogeneous transition models.  We denote them %
VI-short, VI-hom, and VI-short-hom.

\myparagraph{Local control} 
The local control module takes in an observation vector from the vision module, an action-value vector from the planner, and outputs a new action.  We experiment with a hand-built state machine policy (SM) and learned policies, based either on the 3-vector vision output (LSTM) or the richer feature output (f-LSTM).

An alternative to a local control policy is to update the map based on the observations and replan.  We denote that case as VI-repl for the planning module, although it can also be thought of as a local-control strategy.

\subsection{Training Regimes}
We consider two main approaches for training modules: independent, conventional model learning that optimizes for model accuracy; and joint learning that optimizes for the final policy objective end-to-end, using DAN. %
Both approaches use trajectories from $10,000$ random environments for training, 5 trajectories from each environment. 

For independent model learning we use appropriately labelled data tuples: $(o_t, v_t)_i$ for a vision module, where $o_t$ is an image and $v_t$ is a local observation vector;  $(s_t, v_t, M)_i$ tuples for an observation model where $M$ is a map; $(s_{t}, a_t, s_{t+1}, M)_i$ tuples for a transition model; and $(s_t, a_t, r_t, M)_i$ tuples for a reward model. The training loss is defined by the appropriate negative log-likelihood of the output.

For end-to-end learning we use expert demonstrations in the form of observation-action trajectories and a map, $(o_0, a_0, o_1, a_1, ..., o_t, a_t, M)_i$. Expert trajectories are obtained by near-optimal, clairvoyant SARSOP policies that have full knowledge of the underlying environment model. The expert has access to more information than the robot: it uses ground-truth vector observations in Task C and D, and it uses the underlying location of the robot to build a map in Task D. The training loss is the imitation learning objective, \ie, the negative log-likelihood of the expert action for each step along the trajectories. To account for critical but rare events in Task D, we inflated the loss by a factor of 10 when the expert policy had to avoid an unexpected obstacle. 
We train DANs in a curriculum of increasingly difficult task variants; and pre-train the vision module independently. Details are in Appendix~\ref{sec:implementation}.%

\section{Experimental results}

Quantitative results on different combinations of modules, training strategies and tasks are shown in \tabref{tab:main_results}.  Each row corresponds to one experiment. The first two columns characterize the task.  The third column describes the architecture and training regime: it lists some subset of the modules V (vision), F (filter), P (planning), and C (local control);  the notation DAN(X) means that the modules inside the parentheses are jointly trained to optimize task performance; modules not inside a DAN grouping are either fixed, or have models trained to optimize a predictive objective.  The next columns indicate the particular implementation of the component (a detailed list is in Appendix~\ref{sec:module_details}). Finally, we report the performance of the trained system: the percentage of trials the goal is reached~(within 238 steps), the average number of time steps to the goal~(only including the successful trials), the percentage of trials that involved one or more collisions, and the total discounted reward averaged over trials. Evaluation is in $500$ randomly generated environments that were not seen during training. 

A video is available at \href{https://youtu.be/4jcYlTSJF4Y}{https://youtu.be/4jcYlTSJF4Y}.

In the following we walk through the experiments, pointing out the most salient results and illustrating a number of ways in which the DAN approach facilitates robot system design. %
A summary of our findings is in~\secref{sec:summary}.

\subsection{Task A: Discrete MDP}
We begin with a value-iteration algorithm and model representation that are complete for this task (VI*).  Whether we train the models independently (A1) or via DAN (A2), the system performs nearly optimally.  

We then consider reducing the horizon of value iteration to 25.  In this scenario, using independently trained models (A3) performs poorly, because most problem instances require more than 25 steps.  Using DAN training (A4), we recover near-optimal performance. As in the Puddle-MDP, DAN learns models that compensate for the weakness of the algorithm: we find that the learned transition model predicts that actions move farther than they actually do.  %

Learning a transition model that can be different for each state in the domain is costly in space, computation time, and training examples.  In many domains, a {\em spatially homogeneous} model, which predicts transitions relative to the robot's current location, will suffice;  such models have a relatively small number of parameters and are insensitive to the size of the domain. In our domain, the dynamics are similar in all parts of the space, but do in fact differ locally because of the presence of walls.  Experiment A5 illustrates that learning the maximum-likelihood homogeneous transition model and planning with it as if it were exactly correct yields very poor performance.
However, applying DAN training to this model (A6) recovers near-optimal performance.  Examining the learned models, we see that the penalty for collisions is inflated, causing the robot to select actions that will keep it farther away from obstacles. 
\pk{evidence}

In experiments A7 and A8 we both plan with a shortened horizon and learn a homogeneous transition model, with the now expected poor performance of independently trained models and significant improvement by DAN.  
We find that, among other things, DAN discovers that it is rarely necessary to take two turn actions in a row; instead, the learned transition model combines the effects of a turn and subsequent move, thus shortening the necessary planning horizon.
\pk{evidence}

\newcommand{\dan}[1]{DAN(#1)}
\newcommand{\smallgap}{\abovestrut{0.20in}}
\newcommand{\largegap}{\midrule}

\begin{table*}[t]
\vspace{-8pt} %
\scalebox{0.69}{
\begin{tabular}{ccc|c|cccc|cccc}
\toprule
\vspace{-1pt} %
& \textbf{Map} & \textbf{Observation} & \textbf{Architecture} & \textbf{Vision} & \textbf{Filter} & \textbf{Planner} & \textbf{Control}  & \textbf{Success} & \textbf{Time} & \textbf{Collision} & \textbf{Reward}\\
& \textbf{} & \textbf{} & \textbf{} & \textbf{V} & \textbf{F} & \textbf{P} & \textbf{C} & \textbf{rate} & \textbf{steps} & \textbf{rate} & \textbf{}\\
\midrule[0.08em] %
A1 & full & state & P  &   &   & VI* &  & \textbf{100\%} & 32.80 & 0\% & \textbf{12.693} \\
A2 & full & state & \dan{P} &   &   & VI* &   & \textbf{100\%} & 32.80 & 0\% & 12.692 \\
\smallgap
A3 & full & state & P  &   &   & VI-short &   & 59.0\% & 19.50 & 0\% & 5.459 \\
A4 & full & state & \dan{P} &   &   & VI-short &   & \textbf{97.8\%} & 34.13 & 0\% & \textbf{12.049} \\
\smallgap
A5 & full & state & P &  &   & VI-hom &  & 34.2\% & 13.88 & 65.8\% & -554.3 \\
A6 & full & state & \dan{P} &  &   & VI-hom &   & \textbf{100\%} & 32.83 & 0\% & \textbf{12.686} \\
\smallgap
A7 & full & state & P &   &   & VI-short-hom &   & 33.2\% & 13.59 & 60.4\% & -510.8 \\
A8 & full & state & \dan{P} &  &   & VI-short-hom &  & \textbf{95.0\%} & 35.82 & 0\% &  \textbf{11.241} \\

\largegap
B1 & full & vector & P &   &   &  SARSOP &   &  \textbf{100\%} & 36.12 & 0\% & \textbf{11.97} \\
\smallgap
B2 & full & vector & F + P  &   &  HF-ml & VI &  & 94.4\% & 36.13 & 11.4\% & 2.83 \\
B3 & full & vector & F + P &   &  HF-bel & VI &   &  63.8\% & 36.81 & 0\% & 4.27 \\
B4 & full & vector & \dan{F + P}  &   &  HF & VI &  & \textbf{99.6\%} & 36.49 & 0\% & \textbf{11.85} \\

\largegap
C1 & full & image & V+P & CNN &  & SARSOP &  & \textbf{98.4\%} & 38.07 & 0.6\% & \textbf{11.21} \\
\smallgap
C2 & full & image & V+F+P & CNN & HF-ml & VI &  & 95.2\% & 37.63 & 13.2\% & 3.10 \\
C3 & full & image & V+F+P & CNN & HF-bel & VI &  & 61.8\% & 36.18 & 0.4\% & 3.89 \\
C4 & full & image & V+DAN(F+P) & CNN & HF & VI & & 97.4\% & 40.04 & 0.0\% & 10.80 \\
C5 & full & image & DAN(V+F+P) & CNN & HF & VI & & 98.8\% & 36.30 & 0.8\% & 11.03 \\
C6 & full & image & DAN(V+F+P) & CNN-f & f-HF & VI & & \textbf{99.0\%} & 34.35 & 0.4\% & \textbf{12.09} \\

\largegap
D1 & partial & image & V+F+P & CNN &  HF-ml &  VI-repl &   & \textbf{ 89.6\%} & 39.81 & 7.0\% & -29.40 \\
D2 & partial & image & V+F+P & CNN &  HF-bel &  VI-repl &   & 58.4\% & 36.47 & 5.4\% & -15.64 \\
D3 & partial & image & V+F+P+C & CNN &  HF-ml &  VI &  SM  & 41.4\% & 29.96 & 1.8\% & -11.18 \\
D4 & partial & image & V+F+P+C & CNN &  HF-bel &  VI &  SM  & 40.0\% & 29.72 & 0.8\% & \textbf{-5.42} \\
D5 & partial & image & V+F+P+C & CNN &  HF-bel &  VI &  LSTM  &  66.0\% & 43.18 & 4.4\% & -15.87 \\
D6 & partial & image & V+F+P+C* & CNN &  HF-bel &  VI &  LSTM  & 76.6\% & 40.53 & 3.0\% & -12.14 \\

\smallgap
D7 & partial & image & V+\dan{F+P}+C & CNN &  HF &  VI &  LSTM  & 
96.4\% & 45.17 & 3.6\% & 8.11 \\
D8 & partial & image & V+\dan{F+P}+C* & CNN &  HF &  VI &  LSTM  & 
70.2\% & 42.34 & 3.2\% & 3.60 \\
D9 & partial & image & V+\dan{F+P+C} & CNN &  HF &  VI &  LSTM & 98.6\% & 40.55 & 4.0\% & 4.12 \\
D10 & partial & image & \dan{V+F+P+C} & CNN &  HF &  VI &  LSTM  &  99.4\% & 39.60 & 0.2\% & 11.09 \\
D11 & partial & image & \dan{V+F+P+C} & CNN-f &  f-HF &  VI &  f-LSTM  &  \textbf{99.8\%} & 38.05 & 0.8\% & \textbf{11.43} \\
\bottomrule
\end{tabular}
}
{\caption{Main results}\label{tab:main_results}\vspace{-3pt}}
\end{table*}

\subsection{Task B: Discrete POMDP}
In this setting the robot's location is not observed. Instead it receives local (but noise-less) length-3 observation vectors.
The task can be perfectly modelled by a parameterized POMDP, conditioned on the map.  %
We begin by using a model-class that can represent the  domain exactly, learn models independently, and apply SARSOP~\cite{kurniawati2008sarsop}, a near-optimal POMDP solver to the learned models, which yields near-optimal behavior (B1).  However, solving a POMDP  is expensive (in our case sometimes over $5$ minutes) and it tends to grow doubly exponentially with the horizon. We explore lower-cost, decomposed solutions.

We consider a modular system that decomposes partially observable planning into state estimation with a histogram filter, HF, plus fully observable planning with value iteration, VI. 
The HF depends on observation and transition models, and VI on transition and reward models.  We consider two different fixed interfaces between the modules.  In B2, we extract the most likely state and perform VI as if the robot were certain it was in that state.  The robot tends to reach the goal, but at a cost of frequent collisions.  In B3, we initialize the state-occupancy distribution for VI to be the current belief: this corresponds to the QMDP approximation~\cite{littman1995learning}, which accounts for current uncertainty, but ignores future uncertainty. We first train the models independently (B3) and find that the system performs poorly.  It often gets stuck, oscillating or taking a {\em stay} action, because the approximation assumes that state uncertainty will be dispelled after any action, which is not true in this domain.

Using DAN training to jointly optimize models both in the filter and the planner, directly for task performance, can compensate for the strong approximation in the decoupled system, resulting in near-optimal behavior (B4).  
Because the filter and planner modules are trained jointly, the system is free to adapt the models in a way to optimize the combined system.  We observe that the learned reward model has a large cost for the {\em stay} action, and the learned transition model differs from the true model in ways that break symmetry, hence gathering information and preventing oscillatory behavior.
\pk{evidence}

\subsection{Task C: POMDP with Image Input}
In this setting, instead of three-bit noise-free observations, the robot receives images.  A full POMDP model that operates in the space of images is intractable.  Hence, we add a vision module, in the form of a CNN.  In most experiments in this section, its output is a vector of 3 binary classifications with the same expected semantics as inputs in Task B.

We begin with a classic architecture, in which the vision module is trained via supervised learning, and the observation, transition, and reward models from B1 are used to generate a near-optimal POMDP policy, mapping sequences of outputs from the vision module into actions.  C1 shows that reasonable performance can be obtained this way, although a large amount of computation is needed to solve for the policy, and there are failures due to imperfect vision.

We attempt to reduce computational complexity by combining the independently trained vision module with both a most-likely-state, and a QMDP approximation, based on independently trained models (C2 and C3). We obtain  performance similar to B2 and B3.  Applying DAN training to just the F and P modules (C4) improves performance. Applying DAN to the whole composition of V, F, and P (C5) further improves performance. In each case, we have relaxed the semantics of the interfaces between modules. In C5, we believe the DAN learns to capture the prediction confidence of vision in the observation model allowing increased robustness. %

Finally, we relax the constraint on the interface between vision and state estimation. Instead of going through a binary classification vector, we allow the DAN to learn an observation model for state estimation directly using features extracted from images. Performance further improves, even beyond C1. In particular, the goal is reached in fewer steps.
This implies that, by removing the interface constraint, the DAN learns to extract more information from images for faster localization---the images, in fact, contain more information than just the occupancy of the three cells in front of the robot.   %

\subsection{Task D: POMDP with Image and Partial Map Input}
In this setting the maps given to the robot no longer perfectly describe the environment: they indicate walls, but not the furniture. Even without the problems of image processing, the partially observable decision-making problem is intractable:  both the robot state and the environment are partially observed, resulting in an $N \mytimes 2^N$-dimensional belief space for $N$ grid cells. It is also not clear how the problem can be decomposed. 

One simple strategy is to try the same architectures as for Task C.  Unmodified, they fail disastrously, as the robot repeatedly attempts to move through unmodeled obstacles. A simple idea is to put newly encountered obstacles into the map---but this solution presupposes that the robot is localized!  D1 and D2 show the results of updating the map as if the robot were at the most likely location and replanning whenever an obstacle is encountered.  There is some success with this approach, but with many collisions.

A better strategy is to add a local controller to avoid obstacles, which would select alternative actions if the nominally commanded action would result in collision.  We begin by using a simple hand-coded strategy in D3 and D4:  it largely avoids collisions, but at the cost of often not reaching the goal.
Next, we replace the fixed local controller with an LSTM network that maps the commanded action and the current inputs to an updated action. In D5, we train the LSTM in isolation, with the same distribution of objects, but in a fully observed setting~(C). In D6, we train the LSTM in a partially observable setting, with the HF and VI modules from system D3, still in isolation~(C*). The learning objective is the same as for training DAN, but only the LSTM is trained.  Performance remains poor.

In experiments D7 through D11, 
we apply the DAN methodology to different groups of subsystems of the full architecture.  We find that allowing the system to adapt all of the models and to choose interfaces between the modules gives the best performance. The overall reward 
very
closely rivals the optimal solution in a completely observable version of the problem, even when challenged by image interpretation and incorrect maps.

\subsection{Unstructured Learning Systems}
We have reported on extensive experiments that illustrate how the DAN learning can improve significantly over classical model-learning.  In addition, we experimented with standard neural-networks on the same tasks, with the same end-to-end objective and training data as for the DAN systems. We used simple combinations of CNN and LSTM components, and performed a  basic search over hyper-parameters. Details are in Appendix~\ref{sec:unstructured_details}. We report the best achieved results for each task in \tabref{tab:model_free_results}, which are generally quite poor.

These results are consistent with those from prior work on learning map-based navigation, but in much simpler settings, \eg, \cite{tamar2016value, karkus2017qmdp}. 
It is, of course, entirely possible that a greater investment in the search over neural network architectures and hyper-parameters, as well as the acquisition of more training data, would result in significantly better performance.  

The important message of the DAN approach is that it allows robotics experts to use their prior understanding of the problem, in the form of model-based algorithms, to effectively structure a learning system that has the appropriate bias. Learning is more efficient, both in terms of the engineer's time to set up the problem, and in terms of robot time to gather data. At the same time flexibility and performance can significantly increase over the traditional model-based learning approach.

\begin{table}[t]
\vspace{-8pt} %
\capbtabbox[1.0\textwidth]{
\scalebox{0.65}{
\begin{tabular}{lcc|cccc}
\toprule
\vspace{-1pt} %
    & \textbf{Map} & \textbf{Observation} & \textbf{Success}  & \textbf{Time} & \textbf{Collision} & \textbf{Reward} \\
    & \textbf{} & \textbf{} & \textbf{rate} & \textbf{steps} & \textbf{rate} & \textbf{} \\
 \midrule[0.08em] %
A & full & state & 92.6\% & 39.79 & 9.2\% & -20.65 \\
B & full & vector & 22.2\% & 50.40 & 1.6\% & -4.86 \\  
C & full & image &  38.2\%  & 74.50 & 82.8\% & -103.59 \\
D & partial & image & 38.4\% & 75.97 & 11.4\% & -8.94 \\
\bottomrule
\end{tabular}
}}
{\caption{Results for unstructured neural-network learning}\label{tab:model_free_results}\vspace{3pt}}
\end{table}

\subsection{Summary}\label{sec:summary}
The experimental results point to several strengths of DAN.

\myparagraph{Compensate for approximate algorithms.} 
In Task A, the state-transition model learns ``macro actions'' to compensate for an approximate planner with short horizon (A3-A4, A7-A8).

\myparagraph{Compensate for model mis-specification.} 
In Task A, the reward model learns inflated collision penalty to compensate for a transition model assumed to be homogeneous incorrectly (A5-A6, A7-A8).

\myparagraph{Compensate for approximate decomposition.} 
In Task B, the reward model learns to penalize the {\em stay}
action to compensate for the approximate decomposition of  partially
observable planning into state estimation and fully observable planning system (B3-B4).
Further, in Task C and Task D, the vision model possibly learns to encode uncertainty in its output, and the observation model of the filter learns to take the uncertainty into account. This improves the overall performance of state estimation (C5-C6, D10-D11).

\myparagraph{Unify model-based and model-free representations.}
In Task D, the model-free LSTM controller module, integrated into and trained jointly with a model-based modular system, learns to select actions that compensate for the incompleteness of the map based on perceptual input (D5-D10).

Together the experimental results suggest that through end-to-end learning, DAN
compensates for imperfections in models, algorithms, and system decomposition.
Further, DAN substantially outperforms unstructured neural networks on the evaluation tasks, 
demonstrating the benefits of structured priors through models, algorithms, and system decomposition. %

\section{Conclusion}
DAN  represents a first step towards a general methodology for designing and
implementing robot learning  systems.  It combines well-understood structures---models, algorithms, interfaces---with end-to-end learning. 
This enables DAN  to achieve strong performance in the presence of modeling
errors, while  using  only limited   training  data. 
Our case study suggests that DAN scales up to moderately complex robotic systems
involving multiple common components. %

DAN adapts the models to compensate for various imperfections in manually specified
structures. %
The improved task performance comes at the cost of reduced model reusability. This is, however, acceptable, when no alternative better structures are known.

Scalability is a major challenge in applying the DAN
methodology in practice. A complex systems results in a large network.
Further, the optimization landscape of DANs may be more
challenging than that of over-parameterized, unstructured neural
networks.  To alleviate the difficulty of training, one may
add  ``skip'' connections or improve  credit assignment~\cite{weber2019credit}.
Another interesting direction is to combine classical local
model learning with end-to-end learning, possibly in a
training curriculum. Related ideas in reinforcement learning are promising~\cite{jaderberg2016reinforcement}.

\section*{Acknowledgements}

\begin{spacing}{0.75}
{\footnotesize
We thank Ngiaw Ting An Ian and Aseem Saxena for help with the experimental infrastructure. 
This work is supported, in part, by the Singapore MoE AcRF grant 2016-T2-2-068; ONR Global and AFRL grant N62909-18-1-2023; 
NSF grants 1523767 and 1723381; AFOSR grant FA9550-17-1-0165; ONR grant N00014-18-1-2847; Honda Research; and the MIT-Sensetime Alliance on AI. PK is supported by the NUS Graduate School for Integrative Sciences and Engineering Scholarship.
}
\end{spacing}

\newpage  

\balance    %

\bibliographystyle{plainnat}
\bibliography{references}

\begin{thebibliography}{51}
\providecommand{\natexlab}[1]{#1}
\providecommand{\url}[1]{\texttt{#1}}
\expandafter\ifx\csname urlstyle\endcsname\relax
  \providecommand{\doi}[1]{doi: #1}\else
  \providecommand{\doi}{doi: \begingroup \urlstyle{rm}\Url}\fi

\bibitem[Agrawal et~al.(2016)Agrawal, Nair, Abbeel, Malik, and
  Levine]{agrawal2016learning}
Pulkit Agrawal, Ashvin~V Nair, Pieter Abbeel, Jitendra Malik, and Sergey
  Levine.
\newblock Learning to poke by poking: Experiential learning of intuitive
  physics.
\newblock In \emph{Advances in Neural Information Processing Systems}, pages
  5074--5082, 2016.

\bibitem[Ajay et~al.(2018)Ajay, Wu, Fazeli, Bauza, Kaelbling, Tenenbaum, and
  Rodriguez]{ajay2018augmenting}
Anurag Ajay, Jiajun Wu, Nima Fazeli, Maria Bauza, Leslie~P Kaelbling, Joshua~B
  Tenenbaum, and Alberto Rodriguez.
\newblock Augmenting physical simulators with stochastic neural networks: Case
  study of planar pushing and bouncing.
\newblock In \emph{International Conference on Intelligent Robots and Systems},
  pages 3066--3073, 2018.

\bibitem[Amos and Kolter(2017)]{amos2017optnet}
Brandon Amos and J~Zico Kolter.
\newblock Optnet: Differentiable optimization as a layer in neural networks.
\newblock In \emph{International Conference on Machine Learning}, pages
  136--145, 2017.

\bibitem[Amos et~al.(2018)Amos, Jimenez, Sacks, Boots, and
  Kolter]{amos2018differentiable}
Brandon Amos, Ivan Jimenez, Jacob Sacks, Byron Boots, and J~Zico Kolter.
\newblock Differentiable {MPC} for end-to-end planning and control.
\newblock In \emph{Advances in Neural Information Processing Systems}, pages
  8299--8310, 2018.

\bibitem[Anderson et~al.(2018)Anderson, Chang, Chaplot, Dosovitskiy, Gupta,
  Koltun, Kosecka, Malik, Mottaghi, Savva, et~al.]{anderson2018evaluation}
Peter Anderson, Angel Chang, Devendra~Singh Chaplot, Alexey Dosovitskiy,
  Saurabh Gupta, Vladlen Koltun, Jana Kosecka, Jitendra Malik, Roozbeh
  Mottaghi, Manolis Savva, et~al.
\newblock On evaluation of embodied navigation agents.
\newblock \emph{arXiv preprint arXiv:1807.06757}, 2018.

\bibitem[Andrychowicz et~al.(2018)Andrychowicz, Baker, Chociej, Jozefowicz,
  McGrew, Pachocki, Petron, Plappert, Powell, Ray, et~al.]{And18}
Marcin Andrychowicz, Bowen Baker, Maciek Chociej, Rafal Jozefowicz, Bob McGrew,
  Jakub Pachocki, Arthur Petron, Matthias Plappert, Glenn Powell, Alex Ray,
  et~al.
\newblock Learning dexterous in-hand manipulation.
\newblock \emph{arXiv preprint arXiv:1808.00177}, 2018.

\bibitem[Bansal et~al.(2017)Bansal, Calandra, Xiao, Levine, and
  Tomiin]{bansal2017goal}
Somil Bansal, Roberto Calandra, Ted Xiao, Sergey Levine, and Claire~J Tomiin.
\newblock Goal-driven dynamics learning via bayesian optimization.
\newblock In \emph{IEEE 56th Annual Conference on Decision and Control}, pages
  5168--5173, 2017.

\bibitem[Bellman(2013)]{bellman2013dynamic}
Richard Bellman.
\newblock \emph{Dynamic programming}.
\newblock Courier Corporation, 2013.

\bibitem[Bloom(1970)]{bloom1970space}
Burton~H Bloom.
\newblock Space/time trade-offs in hash coding with allowable errors.
\newblock \emph{Communications of the ACM}, 13\penalty0 (7):\penalty0 422--426,
  1970.

\bibitem[Chen et~al.(2018)Chen, Rubanova, Bettencourt, and
  Duvenaud]{chen2018neural}
Tian~Qi Chen, Yulia Rubanova, Jesse Bettencourt, and David~K Duvenaud.
\newblock Neural ordinary differential equations.
\newblock In \emph{Advances in Neural Information Processing Systems}, pages
  6572--6583, 2018.

\bibitem[Cutler et~al.(2015)Cutler, Walsh, and How]{cutler2015real}
Mark Cutler, Thomas~J Walsh, and Jonathan~P How.
\newblock Real-world reinforcement learning via multifidelity simulators.
\newblock \emph{IEEE Transactions on Robotics}, 31\penalty0 (3):\penalty0
  655--671, 2015.

\bibitem[Dasgupta et~al.(2018)Dasgupta, Sheehan, Stevens, and
  Navlakha]{dasgupta2018neural}
Sanjoy Dasgupta, Timothy~C Sheehan, Charles~F Stevens, and Saket Navlakha.
\newblock A neural data structure for novelty detection.
\newblock \emph{Proceedings of the National Academy of Sciences}, 115\penalty0
  (51):\penalty0 13093--13098, 2018.

\bibitem[Deisenroth and Rasmussen(2011)]{deisenroth2011pilco}
Marc Deisenroth and Carl~E Rasmussen.
\newblock {PILCO}: A model-based and data-efficient approach to policy search.
\newblock In \emph{International Conference on Machine Learning}, pages
  465--472, 2011.

\bibitem[DeSouza and Kak(2002)]{desouza2002vision}
Guilherme~N DeSouza and Avinash~C Kak.
\newblock Vision for mobile robot navigation: A survey.
\newblock \emph{IEEE transactions on pattern analysis and machine
  intelligence}, 24\penalty0 (2):\penalty0 237--267, 2002.

\bibitem[Donti et~al.(2017)Donti, Amos, and Kolter]{donti2017task}
Priya Donti, Brandon Amos, and J~Zico Kolter.
\newblock Task-based end-to-end model learning in stochastic optimization.
\newblock In \emph{Advances in Neural Information Processing Systems}, pages
  5484--5494, 2017.

\bibitem[Doshi-Velez et~al.(2012)Doshi-Velez, Pineau, and
  Roy]{doshi2012reinforcement}
Finale Doshi-Velez, Joelle Pineau, and Nicholas Roy.
\newblock Reinforcement learning with limited reinforcement: Using bayes risk
  for active learning in pomdps.
\newblock \emph{Artificial Intelligence}, 187:\penalty0 115--132, 2012.

\bibitem[Farahmand(2018)]{farahmand2018iterative}
Amir-Massoud Farahmand.
\newblock Iterative value-aware model learning.
\newblock In \emph{Advances in Neural Information Processing Systems}, pages
  9090--9101, 2018.

\bibitem[Farquhar et~al.(2018)Farquhar, Rockt{\"a}schel, Igl, and
  Whiteson]{farquhar2017treeqn}
Gregory Farquhar, Tim Rockt{\"a}schel, Maximilian Igl, and Shimon Whiteson.
\newblock Tree{QN} and {AT}ree{C}: Differentiable tree planning for deep
  reinforcement learning.
\newblock In \emph{Proceedings of the International Conference on Learning
  Representations (ICLR)}, 2018.

\bibitem[Guez et~al.(2018)Guez, Weber, Antonoglou, Simonyan, Vinyals, Wierstra,
  Munos, and Silver]{guez2018learning}
Arthur Guez, Th{\'e}ophane Weber, Ioannis Antonoglou, Karen Simonyan, Oriol
  Vinyals, Daan Wierstra, R{\'e}mi Munos, and David Silver.
\newblock Learning to search with {MCTS}nets.
\newblock In \emph{International Conference on Machine Learning}, pages
  1822--1831, 2018.

\bibitem[Haarnoja et~al.(2016)Haarnoja, Ajay, Levine, and
  Abbeel]{haarnoja2016backprop}
Tuomas Haarnoja, Anurag Ajay, Sergey Levine, and Pieter Abbeel.
\newblock Backprop {KF}: Learning discriminative deterministic state
  estimators.
\newblock In \emph{Advances in Neural Information Processing Systems}, pages
  4376--4384, 2016.

\bibitem[Hochreiter and Schmidhuber(1997)]{hochreiter1997long}
Sepp Hochreiter and J{\"u}rgen Schmidhuber.
\newblock Long short-term memory.
\newblock \emph{Neural Computation}, 9\penalty0 (8):\penalty0 1735--1780, 1997.

\bibitem[Jaderberg et~al.(2016)Jaderberg, Mnih, Czarnecki, Schaul, Leibo,
  Silver, and Kavukcuoglu]{jaderberg2016reinforcement}
Max Jaderberg, Volodymyr Mnih, Wojciech~Marian Czarnecki, Tom Schaul, Joel~Z
  Leibo, David Silver, and Koray Kavukcuoglu.
\newblock Reinforcement learning with unsupervised auxiliary tasks.
\newblock \emph{arXiv preprint arXiv:1611.05397}, 2016.

\bibitem[Jiang et~al.(2015)Jiang, Kulesza, Singh, and
  Lewis]{jiang2015dependence}
Nan Jiang, Alex Kulesza, Satinder Singh, and Richard Lewis.
\newblock The dependence of effective planning horizon on model accuracy.
\newblock In \emph{Proceedings of the International Conference on Autonomous
  Agents and Multiagent Systems}, pages 1181--1189, 2015.

\bibitem[Johannink et~al.(2018)Johannink, Bahl, Nair, Luo, Kumar, Loskyll,
  Ojea, Solowjow, and Levine]{johannink2018residual}
Tobias Johannink, Shikhar Bahl, Ashvin Nair, Jianlan Luo, Avinash Kumar,
  Matthias Loskyll, Juan~Aparicio Ojea, Eugen Solowjow, and Sergey Levine.
\newblock Residual reinforcement learning for robot control.
\newblock \emph{arXiv preprint arXiv:1812.03201}, 2018.

\bibitem[Jonschkowski and Brock(2016)]{jonschkowski2016}
Rico Jonschkowski and Oliver Brock.
\newblock End-to-end learnable histogram filters.
\newblock In \emph{NeurIPS Workshop on Deep Learning for Action and
  Interaction}, 2016.

\bibitem[Jonschkowski et~al.(2018)Jonschkowski, Rastogi, and
  Brock]{jonschkowski2018differentiable}
Rico Jonschkowski, Divyam Rastogi, and Oliver Brock.
\newblock Differentiable particle filters: End-to-end learning with algorithmic
  priors.
\newblock \emph{Proceedings of Robotics: Science and Systems}, 2018.

\bibitem[Karkus et~al.(2017)Karkus, Hsu, and Lee]{karkus2017qmdp}
Peter Karkus, David Hsu, and Wee~Sun Lee.
\newblock {QMDP}-net: Deep learning for planning under partial observability.
\newblock In \emph{Advances in Neural Information Processing Systems}, pages
  4697--4707, 2017.

\bibitem[Karkus et~al.(2018)Karkus, Hsu, and Lee]{karkus2018particle}
Peter Karkus, David Hsu, and Wee~Sun Lee.
\newblock Particle filter networks with application to visual localization.
\newblock In \emph{Proceedings of the Conference on Robot Learning}, pages
  169--178, 2018.

\bibitem[Kloss and Bohg(2019)]{kloss2019on}
Alina Kloss and Jeannette Bohg.
\newblock On learning heteroscedastic noise models within differentiable bayes
  filters, 2019.
\newblock URL \url{https://openreview.net/forum?id=BylBns0qtX}.

\bibitem[Krizhevsky et~al.(2012)Krizhevsky, Sutskever, and
  Hinton]{krizhevsky2012imagenet}
Alex Krizhevsky, Ilya Sutskever, and Geoffrey~E Hinton.
\newblock Imagenet classification with deep convolutional neural networks.
\newblock In \emph{Advances in Neural Information Processing Systems}, pages
  1097--1105, 2012.

\bibitem[Kurniawati et~al.(2008)Kurniawati, Hsu, and Lee]{kurniawati2008sarsop}
Hanna Kurniawati, David Hsu, and Wee~Sun Lee.
\newblock {SARSOP}: Efficient point-based {POMDP} planning by approximating
  optimally reachable belief spaces.
\newblock \emph{Proceedings of Robotics: Science and Systems}, 2008.

\bibitem[LeCun et~al.(1998)LeCun, Bottou, Bengio, Haffner,
  et~al.]{lecun1998gradient}
Yann LeCun, L{\'e}on Bottou, Yoshua Bengio, Patrick Haffner, et~al.
\newblock Gradient-based learning applied to document recognition.
\newblock \emph{Proceedings of the {IEEE}}, 86\penalty0 (11):\penalty0
  2278--2324, 1998.

\bibitem[Littman et~al.(1995)Littman, Cassandra, and
  Kaelbling]{littman1995learning}
Michael~L Littman, Anthony~R Cassandra, and Leslie~P Kaelbling.
\newblock Learning policies for partially observable environments: Scaling up.
\newblock In \emph{International Conference on Machine Learning}, pages
  362--370, 1995.

\bibitem[Mahler et~al.(2017)Mahler, Liang, Niyaz, Laskey, Doan, Liu, Ojea, and
  Goldberg]{mahler2017dex}
Jeffrey Mahler, Jacky Liang, Sherdil Niyaz, Michael Laskey, Richard Doan, Xinyu
  Liu, Juan~Aparicio Ojea, and Ken Goldberg.
\newblock Dex-net 2.0: Deep learning to plan robust grasps with synthetic point
  clouds and analytic grasp metrics.
\newblock \emph{Proceedings of Robotics: Science and Systems}, 2017.

\bibitem[Mirowski et~al.(2016)Mirowski, Pascanu, Viola, Soyer, Ballard, Banino,
  Denil, Goroshin, Sifre, Kavukcuoglu, et~al.]{mirowski2016learning}
Piotr Mirowski, Razvan Pascanu, Fabio Viola, Hubert Soyer, Andy Ballard, Andrea
  Banino, Misha Denil, Ross Goroshin, Laurent Sifre, Koray Kavukcuoglu, et~al.
\newblock Learning to navigate in complex environments.
\newblock \emph{arXiv preprint arXiv:1611.03673}, 2016.

\bibitem[Nilsson(1984)]{nilsson1984shakey}
Nils~J Nilsson.
\newblock Shakey the robot.
\newblock Technical report, SRI AI Center Menlo Park CA, 1984.

\bibitem[Oh et~al.(2017)Oh, Singh, and Lee]{oh2017value}
Junhyuk Oh, Satinder Singh, and Honglak Lee.
\newblock Value prediction network.
\newblock In \emph{Advances in Neural Information Processing Systems}, pages
  6120--6130, 2017.

\bibitem[Okada et~al.(2017)Okada, Rigazio, and Aoshima]{okada2017path}
Masashi Okada, Luca Rigazio, and Takenobu Aoshima.
\newblock Path integral networks: End-to-end differentiable optimal control.
\newblock \emph{arXiv preprint arXiv:1706.09597}, 2017.

\bibitem[Pereira et~al.(2018)Pereira, Fan, An, and Theodorou]{pereira2018mpc}
Marcus Pereira, David~D Fan, Gabriel~Nakajima An, and Evangelos Theodorou.
\newblock {MPC}-inspired neural network policies for sequential decision
  making.
\newblock \emph{arXiv preprint arXiv:1802.05803}, 2018.

\bibitem[Racani{\`e}re et~al.(2017)Racani{\`e}re, Weber, Reichert, Buesing,
  Guez, Rezende, Badia, Vinyals, Heess, Li, et~al.]{racaniere2017imagination}
S{\'e}bastien Racani{\`e}re, Th{\'e}ophane Weber, David Reichert, Lars Buesing,
  Arthur Guez, Danilo~Jimenez Rezende, Adria~Puigdomenech Badia, Oriol Vinyals,
  Nicolas Heess, Yujia Li, et~al.
\newblock Imagination-augmented agents for deep reinforcement learning.
\newblock In \emph{Advances in Neural Information Processing Systems}, pages
  5690--5701, 2017.

\bibitem[Schulman et~al.(2015)Schulman, Heess, Weber, and
  Abbeel]{schulman2015gradient}
John Schulman, Nicolas Heess, Theophane Weber, and Pieter Abbeel.
\newblock Gradient estimation using stochastic computation graphs.
\newblock In \emph{Advances in Neural Information Processing Systems}, pages
  3528--3536, 2015.

\bibitem[Silver et~al.(2018)Silver, Allen, Tenenbaum, and
  Kaelbling]{silver2018residual}
Tom Silver, Kelsey Allen, Josh Tenenbaum, and Leslie Kaelbling.
\newblock Residual policy learning.
\newblock \emph{arXiv preprint arXiv:1812.06298}, 2018.

\bibitem[Srinivas et~al.(2018)Srinivas, Jabri, Abbeel, Levine, and
  Finn]{srinivas2018universal}
Aravind Srinivas, Allan Jabri, Pieter Abbeel, Sergey Levine, and Chelsea Finn.
\newblock Universal planning networks.
\newblock \emph{arXiv preprint arXiv:1804.00645}, 2018.

\bibitem[Talvitie(2014)]{talvitie2014model}
Erik Talvitie.
\newblock Model regularization for stable sample rollouts.
\newblock In \emph{UAI}, pages 780--789, 2014.

\bibitem[Tamar et~al.(2016)Tamar, Levine, Abbeel, Wu, and
  Thomas]{tamar2016value}
Aviv Tamar, Sergey Levine, Pieter Abbeel, Yi~Wu, and Garrett Thomas.
\newblock Value iteration networks.
\newblock In \emph{Advances in Neural Information Processing Systems}, pages
  2146--2154, 2016.

\bibitem[Thrun et~al.(2005)Thrun, Burgard, and Fox]{thrun2005probabilistic}
Sebastian Thrun, Wolfram Burgard, and Dieter Fox.
\newblock \emph{Probabilistic Robotics}.
\newblock MIT Press, 2005.

\bibitem[Thrun et~al.(2006)Thrun, Montemerlo, Dahlkamp, Stavens, Aron, Diebel,
  Fong, Gale, Halpenny, Hoffmann, et~al.]{thrun2006stanley}
Sebastian Thrun, Mike Montemerlo, Hendrik Dahlkamp, David Stavens, Andrei Aron,
  James Diebel, Philip Fong, John Gale, Morgan Halpenny, Gabriel Hoffmann,
  et~al.
\newblock Stanley: The robot that won the {DARPA} {G}rand {C}hallenge.
\newblock \emph{Journal of field Robotics}, 23\penalty0 (9):\penalty0 661--692,
  2006.

\bibitem[{Unity 3D}()]{unity3d}
{Unity 3D}.
\newblock Game engine.
\newblock URL \url{http://unity3d.com}.

\bibitem[Van Den~Berg et~al.(2011)Van Den~Berg, Abbeel, and
  Goldberg]{van2011lqg}
Jur Van Den~Berg, Pieter Abbeel, and Ken Goldberg.
\newblock {LQG-MP}: Optimized path planning for robots with motion uncertainty
  and imperfect state information.
\newblock \emph{The International Journal of Robotics Research}, 30\penalty0
  (7):\penalty0 895--913, 2011.

\bibitem[Weber et~al.(2019)Weber, Heess, Buesing, and Silver]{weber2019credit}
Th{\'e}ophane Weber, Nicolas Heess, Lars Buesing, and David Silver.
\newblock Credit assignment techniques in stochastic computation graphs.
\newblock \emph{arXiv preprint arXiv:1901.01761}, 2019.

\bibitem[Zhu et~al.(2017)Zhu, Mottaghi, Kolve, Lim, Gupta, Fei-Fei, and
  Farhadi]{zhu2017target}
Yuke Zhu, Roozbeh Mottaghi, Eric Kolve, Joseph~J Lim, Abhinav Gupta,
  Li~Fei-Fei, and Ali Farhadi.
\newblock Target-driven visual navigation in indoor scenes using deep
  reinforcement learning.
\newblock In \emph{International Conference on Robotics and Automation}, pages
  3357--3364, 2017.

\end{thebibliography}

\clearpage

\appendices

\section{Implementation details}
\nobalance
\label{sec:implementation}
We implement DAN modules in Tensorflow. The DAN implementation of HF and VI are based on~\cite{karkus2017qmdp} and \cite{tamar2016value} with the following differences. In VI we use soft-max for performing Bellman updates; the computed Q values are directly used without an additional fully-connected layer; and our transition, reward and observation models have a different form depending on the task variant and experiment setting, as discussed in the paper.

We train DAN systems end-to-end using expert demonstration data. We use backpropagation through time~(BPTT) with gradient clipping.  In the partially observable domains the latent belief in DAN is initialized with the true underlying initial belief; however, we do not have access to ground-truth beliefs after the first step. 

For more complex tasks we found it necessary to train DANs in multiple stages of curriculum. In each stage DAN is trained end-to-end with expert demonstrations, in increasingly difficult task settings. Specifically, in Task B, we start with a training stage where initial beliefs are set to be certain at the true underlying state (while the true state is still not observed after the first step). The second training stage corresponds to the full Task B setting, \ie{}, initial beliefs are uniformly distributed across a random subset of the state space. In Task C we use three training stages. We first replace image inputs with binary observations and follow the two training stages of Task B. We then add the vision module to DAN and train further with image observations. In Task D we follow the same training stages as in Task C (but now map inputs are only partially correct). 

The vision models in Task C and Task D are pre-trained independently with ground-truth binary observation vectors. CNN-f is obtained by replacing the last, fully-connected layer of a pre-trained CNN model and initializing the new weights randomly.  We use L2 regularization for the convolutional layers of CNN and CNN-f.

In each training stage we follow an \emph{early stopping with patience} strategy. Specifically, we decay the learning rate by factors of 0.8 in multiple steps. The first decay step is executed when the validation loss is not decreasing for 15 consecutive epochs. Four more decay steps are executed when the validation loss is not decreasing for 5 consecutive epochs. Training terminates when the next decay step would be triggered. The model with lowest validation loss is chosen for evaluation.

We use the RMSProp optimizer with 0.9 decay rate. %
The batch size is set to 100. The initial learning rate is set to 0.001. In Task B and Task C we truncate BPTT at 4 time steps. In Task D the initial learning rate is 0.0005 and we truncate BPPT at 8 time steps. Longer sequences did not improve performance.

We use NVidia GTX1080Ti GPUs. Training required approximately 1--8 GB GPU memory and 6--48 hours depending on the task and training setting.

\section{Detailed Module Descriptions}
\label{sec:module_details}

\subsection*{Vision}
\begin{enumerate}
    \item CNN.  A convolutional neural network with a length-3 binary vector output that indicates the presence of walls and objects in the 3 grid squares in front of the robot.  This output is compatible with the observations provided as input in Task B. The CNN has 3 convolutional layers (3-3-1 kernels and 128-128-1 filters) connected to a fully connected layer that maps to the output vector.
    \item CNN-f.  A similar network as CNN, but with 16-dimensional feature vector outputs that have no pre-determined semantics. %
\end{enumerate}
\subsection*{Filtering}
\begin{enumerate}
    \item HF-bel. Histogram filter that takes in 3-dimensional observation vectors, the last action of the robot, and the previous belief distribution represented by a 1444-dimensional vector. The output is the updated belief distribution.  At the beginning of a trajectory the belief is initialized to the true initial belief.  The transition model in the DAN implementation is homogeneous, in the independently learned implementation it is heterogeneous. 
    \item HF-ml.  The same histogram filter as above, but with the output transformed into a one-hot vector representing the most likely state.
    \item HF. The same histogram filter as HF-bel but when used as a DAN, the output is only constrained to be a distribution, \ie, the vector is normalized, but it is not directly enforced to represent the true belief.
    \item f-HF. The same filter as HF with a different observation model that takes in 16 dimensional observation features with undetermined semantics.
\end{enumerate}
\subsection*{Planning}
\begin{enumerate}
    \item VI. The default setting of value iteration. When implemented as a DAN we use the homogeneous transition model, and H=76. When learned independently, we use the heterogeneous transition model and plan until convergence.
    \item VI*. Value iteration with  a sufficiently long planning horizon~(H=200) to guarantee finding the optimal path and an environment-dependent, heterogeneous transition model. The same setting is used in the independent implementation and in DAN.
    \item VI-short. Value iteration with a short planning horizon~(H=25).
    \item VI-hom. Value iteration with a homogeneous transition model and long planning horizon.
    \item VI-short-hom.  Value iteration with short planning horizon and homogeneous transition model.
    \item VI-repl. In this implementation when an obstacle is detected by the vision module it is added to the map according to the current most likely state. Whenever the map is updated we \emph{replan}, \ie, the $Q$ values are recomputed. 
    \item SARSOP. The SARSOP planner embodies its own Bayes filter. It directly takes in a length-3 binary observation vector and outputs an action. It uses the same transition and reward models as VI, and the same observation model as HF. 
\end{enumerate}
\subsection*{Local control}
\begin{enumerate}
    \item SM. Simple, hand-coded state machine for obstacle avoidance. When the action with the highest value would lead to a collision, it picks one of the turn actions with the higher value. Further, to prevent oscillation, consecutive left-right turns are replaced by a forward action if it does not lead to collision.
    \item LSTM. A parameterized module using LSTM. We concatenate the observation vector, the  action-value vector and the last action vector, and input them to an LSTM layer with 64 hidden units. The output of the LSTM is combined with the input action-values to produce new action values. The action with the highest value is chosen.
    \item f-LSTM. Same as LSTM, but takes in observation vectors with undetermined semantics from the CNN-f vision module.
\end{enumerate}

\section{Unstructured learning systems}\label{sec:unstructured_details}

\balance

The unstructured learning systems all use similar recurrent neural network architectures, which are adapted to the different set of inputs for each task variant. First, features are extracted from the inputs. The map, goal and initial belief inputs are encoded as $19 \mytimes 19 \mytimes x$ images, and processed by $N$ convolutional layers with $3 \mytimes 3$ kernels and 256 convolutional filters. Observations are different for each task. In Task A observations are robot locations. They are represented as one-hot beliefs, and processed together with the map and goal for each time step. In Task B observations are binary vectors. They are processed by two fully connected layers, with 128 and 32 hidden units. In Task C and Task D observations are images. They are processed by the same CNN-f network module as in the model-based systems, which is pre-trained independently with ground-truth binary observation vectors.
The features extracted from all inputs are concatenated, and along with the one-hot encoding of the last action, they are fed to an LSTM layer with $L$ hidden dimensions. The output of the LSTM layer is led to a fully connected layer with $D$ hidden dimensions, which maps to the action output.

We perform a basic search over hyper-parameters independently for each task. We search the number of convolutional layers, $N = \{3, 4, 5\}$; the LSTM hidden dimensions, $L = \{128, 256, 512\}$, and the fully-connected layer hidden dimensions $D = \{512, 1024, 2048\}$. We train in the same imitation learning setup as for DAN systems. We use a learning rate of 0.0005, batch size of 64, BPTT truncated to 8 time steps. We explored varying these training parameters but we found them to have little impact on the final performance.
We report the policy performance for the network architecture that achieved the lowest loss on a validation set. During the hyper-parameter search we terminated training slightly earlier than for DANs; however, after the search we have re-trained with the best hyper-parameter, using the same training schedule and stopping criteria as for DANs.

\end{document}